%% file: main.tex
\documentclass{article}

\usepackage{arxiv}
\usepackage{natbib}
\usepackage{paper}

\usepackage{algorithm}
\usepackage{algorithmic}
\usepackage{enumitem}

\usepackage{float}
\floatstyle{plaintop}
\restylefloat{table}

\usepackage{caption}
\input{prelude}

\title{Free Lunch in the Forest:\\Functionally-Identical Pruning of Boosted Tree Ensembles}

\newif\ifuniqueaffiliation
\uniqueaffiliationfalse

\ifuniqueaffiliation 
\author{}
\else
\usepackage{authblk}

\newcommand\mailto[1]{\href{mailto:#1}{\texttt{#1}}}
\newbox{\orcid}%
\sbox{\orcid}{\includegraphics[scale=0.06]{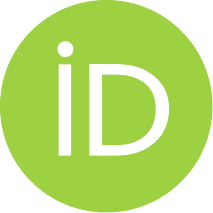}} 
\author[1, 2]{%
    \href{https://orcid.org/0000-0001-8041-5596}{%
        \usebox{\orcid}%
        \hspace{1mm}%
        Youssouf Emine%
        \thanks{\mailto{youssouf.emine@polymtl.ca}}%
    }%
}
\author[1,3,4]{%
    \href{https://orcid.org/0000-0002-9868-4804}{%
        \usebox{\orcid}%
        \hspace{1mm}%
        Alexandre Forel%
        \thanks{\mailto{alexandre.forel@polymtl.ca}}%
    }%
}
\author[3]{%
        \hspace{1mm}%
        Idriss Malek%
        \thanks{\mailto{idriss.malek@polytechnique.edu}}
}%
\author[1,3,4]{%
    \href{https://orcid.org/0000-0001-5183-8485}{%
        \usebox{\orcid}%
        \hspace{1mm}%
        Thibaut Vidal%
        \thanks{\mailto{thibaut.vidal@polymtl.ca}}%
    }%
}
\affil[1]{Department of Mathematics and Industrial Engineering, Polytechnique Montréal}
\affil[2]{Canada Excellence Research Chair in Data-Science for Real-time Decision-Making (CERC)}
\affil[3]{SCALE-AI Chair in Data-Driven Supply Chains}
\affil[4]{Centre Interuniversitaire de Recherche sur les Réseaux d'Entreprise, la Logistique et le Transport
(CIRRELT)}
\fi

\begin{document}

\setlength{\headheight}{24pt}

\maketitle

\import{sections}{0-abstract.tex}
\import{sections}{1-introduction.tex}
\import{sections}{2-problem.tex}
\import{sections}{3-prune.tex}
\import{sections}{4-oracle.tex}
\import{sections}{5-numerical.tex}
\import{sections}{6-literature.tex}
\import{sections}{7-conclusion.tex}
\import{sections}{8-acknowledgements.tex}

\bibliographystyle{aaai25}
\bibliography{main.bib}

\clearpage

\appendix
\import{sections}{a-method.tex}
\import{sections}{b-experiments.tex}

\end{document}

%% file: prelude.tex
\graphicspath{%
    {./images/}%
}

\usetikzlibrary{matrix}

\pgfplotstableset{
    percentage/.style={
        multiply by=100,
        postproc cell content/.append code={%
            \pgfkeysalso{@cell content/.add={}{\%}}%
        },
        fixed,
    },
    fidelity/.style={
        percentage,
        column name = {\bfseries \faithmetric},
        column type = {r},
    },
    elapsed-time/.style={
        column name = {\bfseries Time (\SI{}{\second})},
        column type = {r},
        fixed,
        fixed zerofill,
        precision=0,
    },
    dataset/.style={
        string type,
        column name = {\bfseries Dataset},
        column type = {l},
        column type/.add = {>{\sf} }{}
    },
    method/.style={
        string type,
        column name = {},
        column type = {l},
        column type/.add = {>{\modelfont} }{}
    },
    ensemble/.style={
        string type,
        column name = {%
            \model{Ensemble}%
        },
        column type = {c},
        column type/.add = {>{\modelfont} }{},
        string replace={ab}{%
            \model{AdaBoost}
        },
        string replace={gb}{%
            \begin{tabular}{c}
                \model{Gradient}%
                \\
                \model{Boosting}%
            \end{tabular}
        },
        string replace={lgbm}{%
            \model{LightGBM}
        },
        string replace={xgb}{%
            \model{XGBoost}
        },
        string replace={rf}{%
            \begin{tabular}{c}
                \model{Random}%
                \\
                \model{Forest}%
            \end{tabular}
        },
    },
    n-active-estimators/.style={
        column name = \(\left\| w \right\|_{0}\),
        column type = {r},
        fixed,
        fixed zerofill,
        precision=0,
    },
    accuracy/.style={
        column type = {r},
        column name = {\bfseries \accuracy},
        percentage,
        fixed zerofill,
        precision=2,
    },
    n-oracle-calls/.style={%
        column type = {r},
        column name = {\(\noracle\)},
        precision=0,
        fixed,
    },
    bold-max/.style={
        postproc cell content/.append code={%
            \pgfkeys{/pgf/fpu=false}
            \pgfmathparse{int(mod(\pgfplotstablerow,4))}%
            \ifnum\pgfmathresult=0%
                \pgfkeysalso{
                    @cell content/.add={%
                        \mathversion{tabularbold}%
                    }{}
                }
            \fi
        }
    }
}
\pgfplotsset{
    discard if not/.style 2 args={
        x filter/.code={
            \edef\tempa{\thisrow{#1}}
            \edef\tempb{#2}
            \ifx\tempa\tempb
            \else
                \def\pgfmathresult{inf}
            \fi
        }
    }
}

\newcommand\quoteIt[1]{``\textit{#1}''}
\newcommand\modelfont{\tt}
\newcommand\model[1]{{\modelfont #1}}
\newcommand\package[1]{\texttt{#1}}

\newcommand\class{c}

\newcommand\finiteset{\mathcal{D}}

\newcommand\newpoints{\mathcal{S} }

\newcommand\faithmetric{\textsc{Fi}}
\newcommand\accuracy{\textsc{Acc}}
\newcommand\noracle{K}

\newcommand\treeroot{\mathbf{root} }
\newcommand\treeleft{\mathbf{left} }
\newcommand\treeright{\mathbf{right} }
\newcommand\treedepth{\mathbf{depth} }
\newcommand\node{v}
\newcommand\nodes{\mathcal{V} }
\newcommand\leaves{\mathcal{L} }
\newcommand\flowvar{\mathrm{f} }

\def\fiperepo{%
    \url{https://www.github.com/eminyous/fipe}%
}
\def\experimentsrepo{%
    \url{https://www.github.com/eminyous/fipe-experiments}%
}

\NewDocumentCommand{\showDatasets}{m}{%
    \pgfplotstableread[
        col sep=comma,
        header=true,
        comment chars={\#},
        columns/dataset/.style={string type}
    ]{#1}\loadedtable
    \pgfplotstabletypeset[
        fixed,
        every head row/.style={%
            before row=\toprule,
            after row=\toprule
        },
        every last row/.style={after row=\bottomrule},
        columns = {
            dataset,
            n-samples,
            n-features,
            n-numeric,
            n-binary,
            n-classes
        },
        columns/dataset/.style = {dataset},
        columns/n-samples/.style = {
            column name = {\(n\)},
            column type = {r}
        },
        columns/n-features/.style = {
            column name = {\(p\)},
            column type = {r}
        },
        columns/n-numeric/.style = {
            column name = {\(p_N\)},
            column type = {r}
        },
        columns/n-binary/.style = {
            column name = {\(p_B\)},
            column type = {r}
        },
        columns/n-categorical/.style = {
            column name = {\(p_C\)},
            column type = {r}
        },
        columns/n-classes/.style = {
            column name = {\(C\)},
            column type = {r}
        },
        multicolumn names,
        sort,
        sort key = n-samples,
    ]{\loadedtable}
}

\NewDocumentCommand\showLzeroVSLoneSingle{m}{%
    \pgfplotstableread[
        col sep=comma,
        header=true,
        comment chars={\#},
        columns/dataset/.style={string type}
    ]{#1}\loadedtable

    \xdef\thiscolumns{dataset}
    \foreach \N in {0,1} {
        \xdef\thiscolumns{\thiscolumns, n-active-estimators-l\N}
        \xdef\thiscolumns{\thiscolumns, n-oracle-calls-l\N}
        \xdef\thiscolumns{\thiscolumns, fidelity-l\N}
        \xdef\thiscolumns{\thiscolumns, elapsed-time-l\N}
    }
    \xdef\thiscolumns{\thiscolumns, accuracy-l0}

    \pgfplotsinvokeforeach{0,1}{
        \pgfplotstableset{
            columns/n-active-estimators-l##1/.style={n-active-estimators},
            columns/n-oracle-calls-l##1/.style={n-oracle-calls},
            columns/fidelity-l##1/.style={fidelity},
            columns/elapsed-time-l##1/.style={elapsed-time},
            columns/accuracy-l##1/.style={accuracy},
        }
    }

    \pgfplotstabletypeset[
        columns/.expand once=\thiscolumns,
        columns/dataset/.style={dataset},
        every head row/.style={%
            before row={
                \toprule &%
                \multicolumn{4}{c}{\model{FIPE}\(-\left\| \cdot \right\|_{0}\)} &%
                \multicolumn{4}{c}{\model{FIPE}\(-\left\| \cdot \right\|_{1}\)} &%
                \\
                \cmidrule(ll){2-5}%
                \cmidrule(ll){6-9}%
            },
            after row=\midrule%
        },
        every last row/.style={after row=\bottomrule},
        multicolumn names,
        sort,
        sort cmp={string <},
        sort key=dataset,
    ]{\loadedtable}
}

\NewDocumentCommand\showBaseline{m}{
    \pgfplotstableread[
        col sep=comma,
        header=true,
        comment chars={\#},
        columns/method/.style={string type}
    ]{#1}\loadedtable
    \pgfplotstabletypeset[
        columns={
            n-fitted-estimators,
            n-active-estimators,   
            method,
            fidelity,
            accuracy-after-pruning
        },
        columns/method/.style={
            method
        },
        columns/n-fitted-estimators/.style={%
            column name = \(M\),
            column type = {r},
            assign cell content/.code={%
                \pgfmathparse{int(mod(\pgfplotstablerow,4))}%
                \ifnum\pgfmathresult=0%
                    \pgfkeyssetvalue{/pgfplots/table/@cell content}{%
                        \multirow{4}{*}{####1}
                    }%
                \else
                    \pgfkeyssetvalue{/pgfplots/table/@cell content}{}%
                \fi
            },
        },
        columns/n-active-estimators/.style={%
            column name = \(\left\|w\right\|_{0}\),
            column type = {r},
            assign cell content/.code={%
                \pgfmathparse{int(mod(\pgfplotstablerow,4))}%
                \ifnum\pgfmathresult=0%
                    \pgfkeyssetvalue{/pgfplots/table/@cell content}{%
                        \multirow{4}{*}{%
                            \pgfmathprintnumber[
                                fixed zerofill,
                                precision=2
                            ]{####1}
                        }
                    }%
                \else
                    \pgfkeyssetvalue{/pgfplots/table/@cell content}{}%
                \fi
            },
        },
        columns/fidelity/.style={
            bold-max,
            fidelity,
            fixed,
            fixed zerofill,
            precision=2
        },
        columns/accuracy-after-pruning/.style={
            bold-max,
            accuracy
        },
        every head row/.style={%
            before row=\toprule,%
            after row=\midrule%
        },
        every last row/.style={after row=\bottomrule},
        every nth row={4}{before row=\cmidrule(ll){1-5}},
        multicolumn names,
    ]{\loadedtable}
}

\pgfplotstableread{csv/datasets.csv}\loadedtable%
\pgfplotstablegetrowsof{\loadedtable}
\pgfmathsetmacro{\NDATASETS}{\pgfplotsretval}
\pgfmathsetmacro{\NSEEDS}{5}
\pgfmathsetmacro{\TRAINSPLIT}{0.8*100}
\pgfmathsetmacro{\TESTSPLIT}{0.2*100}
\pgfmathsetmacro{\LZEROEST}{50}

\def\nLinesPerLongTablePage{44}

\NewDocumentCommand{\showLzeroVSLoneAll}{m m}{
    \pgfplotstableread[
        col sep=comma,
        header=true,
        comment chars={\#},
        columns/enemble/.style={string type}
        columns/dataset/.style={string type}
    ]{#1}\loadedtable 
    \xdef\thiscolumns{ensemble, max-depth, dataset}
    \foreach \N in {0,1} {
        \xdef\thiscolumns{\thiscolumns, n-active-estimators-\N}
        \xdef\thiscolumns{\thiscolumns, n-oracle-calls-\N}
        \xdef\thiscolumns{\thiscolumns, fidelity-\N}
        \xdef\thiscolumns{\thiscolumns, elapsed-time-\N}
    }
    \xdef\thiscolumns{\thiscolumns, accuracy-0}

    \pgfplotsinvokeforeach{0,1}{
        \pgfplotstableset{
            columns/n-active-estimators-##1/.style={
                n-active-estimators
            },
            columns/n-oracle-calls-##1/.style={
                n-oracle-calls
            },
            columns/fidelity-##1/.style={
                fidelity
            },
            columns/elapsed-time-##1/.style={
                elapsed-time
            },
            columns/accuracy-##1/.style={
                accuracy
            },
        }
    }

    \pgfplotstabletypeset[
        columns/.expand once=\thiscolumns,
        columns/dataset/.style={dataset},
        columns/ensemble/.style={
            ensemble,
            assign cell content/.code={%
                \pgfmathtruncatemacro\row{\pgfplotstablerow}%
                \pgfmathparse{int(mod(\row,22))}%
                \ifnum\pgfmathresult=0%
                \pgfkeyssetvalue{/pgfplots/table/@cell content}{%
                    \multirow{22}{*}{####1}
                }%
                \else
                \pgfkeyssetvalue{/pgfplots/table/@cell content}{}%
                \fi%
            },
        },
        columns/max-depth/.style={
            column name = {%
                \texttt{max}%
                \ %
                \texttt{depth}%
            },
            column type = {c},
            fixed,
            fixed zerofill,
            precision=0,
            assign cell content/.code={%
                \pgfmathparse{int(mod(\pgfplotstablerow,11))}%
                \ifnum\pgfmathresult=0%
                \pgfkeyssetvalue{/pgfplots/table/@cell content}{%
                    \multirow{11}{*}{####1}
                }%
                \else
                \pgfkeyssetvalue{/pgfplots/table/@cell content}{}%
                \fi
            },
        },
        every head row/.style={%
            before row={
                #2%
                \\
                \toprule
                &%
                &%
                &%
                \multicolumn{4}{c}{%
                    \model{FIPE}\(-\left\| \cdot \right\|_{0}\)%
                }%
                &%
                \multicolumn{4}{c}{%
                    \model{FIPE}\(-\left\| \cdot \right\|_{1}\)%
                }%
                &%
                \\
                \cmidrule(ll){4-7}%
                \cmidrule(ll){8-11}%
                \endfirsthead%
                \\
                \toprule
                &%
                &%
                &%
                \multicolumn{4}{c}{%
                    \model{FIPE}\(-\left\| \cdot \right\|_{0}\)%
                }%
                &%
                \multicolumn{4}{c}{%
                    \model{FIPE}\(-\left\| \cdot \right\|_{1}\)%
                }%
                \\
                \cmidrule(ll){4-7}%
                \cmidrule(ll){8-11}%
                \model{Ensemble}%
                & \texttt{max} \texttt{depth}%
                & \multicolumn{1}{c}{%
                    \textbf{Dataset}%
                }%
                & \(\left\| w \right\|_{0}\)%
                & \(\noracle\)%
                & \multicolumn{1}{c}{%
                    \textbf{\faithmetric}%
                }%
                & \textbf{Time} (\SI{}{\second})%
                & \(\left\| w \right\|_{0}\)%
                & \(\noracle\)%
                & \multicolumn{1}{c}{%
                    \textbf{\faithmetric}%
                }%
                & \textbf{Time} (\SI{}{\second})%
                & \multicolumn{1}{c}{%
                    \textbf{\accuracy}%
                }%
                \\
                \midrule
                \endhead
            },
            after row=\midrule%
        },
        every last row/.style={after row=\bottomrule},
        every nth row={11}{%
            before row=\cmidrule(ll){2-12}%
        },
        every nth row={22}{%
            before row=\cmidrule(ll){1-12}%
        },
        every nth row={\nLinesPerLongTablePage}{%
            before row={
                \cmidrule(ll){1-12}%
                \pagebreak%
            }
        },%
        begin table=\begin{longtable},
        end table=\end{longtable},
        multicolumn names,
    ]{\loadedtable}
}

\NewDocumentCommand{\showLoneAgg}{m m m m}{%
    \pgfplotstableread[
        col sep=comma,
        header=true,
        comment chars={\#},
        columns/ensemble/.style={string type}
    ]{#1}\loadedtable 

    \xdef\thiscolumns{ensemble, max-depth}
    \foreach \M in {#2,#3,#4} {
        \ifx\M\empty
        \else
            \xdef\thiscolumns{%
                \thiscolumns,%
                prop-\M%
            }
        \fi
    }
    \foreach \M in {#2,#3,#4} {
        \ifx\M\empty
        \else
            \xdef\thiscolumns{%
                \thiscolumns,%
                accuracy-\M%
            }
        \fi
    }

    \pgfplotsinvokeforeach{#2,#3,#4} {
        \ifx##1\empty
        \else
            \pgfplotstableset{
                columns/accuracy-##1/.style={
                    column name = {##1},
                    column type = {r},
                    dec sep align,
                    fixed,
                    fixed zerofill,
                    precision = 2,
                    multiply by=100,
                    postproc cell content/.append code={%
                        \ifnum\pgfplotstablepartno=0%
                        \else%
                            \pgfkeysalso{%
                                @cell content/.add={}{\%}%
                            }%
                        \fi%
                    },%
                },
                columns/prop-##1/.style={
                    column name = {##1},
                    column type = {r},
                    dec sep align,
                    fixed,
                    fixed zerofill,
                    precision = 2,
                    multiply by=100,
                    postproc cell content/.append code={%
                        \ifnum\pgfplotstablepartno=0%
                        \else%
                            \pgfkeysalso{%
                                @cell content/.add={}{\%}%
                            }%
                        \fi%
                    },%
                },%
            }%
        \fi%
    }

    \pgfplotstabletypeset[
        columns/.expand once=\thiscolumns,
        columns/ensemble/.style={
            ensemble,
            column name = {},
            assign cell content/.code={%
                \pgfmathtruncatemacro\row{\pgfplotstablerow}%
                \ifnum\row=0%
                    \pgfkeyssetvalue{/pgfplots/table/@cell content}{%
                        \model{AdaBoost}%
                    }%
                \else
                    \pgfmathparse{int(mod(\row,2))}%
                    \ifnum\pgfmathresult=1%
                    \pgfkeyssetvalue{/pgfplots/table/@cell content}{%
                        \multirow{2}{*}{####1}
                    }%
                    \else
                    \pgfkeyssetvalue{/pgfplots/table/@cell content}{}%
                    \fi%
                \fi
            },
        },
        columns/max-depth/.style={
            column name = {\(M\)},
            column type = {c},
            fixed,
            fixed zerofill,
            precision=0,
            assign cell content/.code={%
                \ifnum\pgfplotstablerow=0%
                    \pgfkeyssetvalue{/pgfplots/table/@cell content}{1}
                \fi
            },
        },
        every head row/.style={
            before row={
                \toprule
                &%
                &%
                \multicolumn{6}{c}{%
                  \(%
                    \displaystyle%
                    \frac{\left\| w \right\|_{0}}{M}%
                  \)%
                }%
                &
                \multicolumn{6}{c}{%
                  \begin{tabular}{c}
                    Test\\
                    Accuracy
                  \end{tabular}
                }%
                \\
                \cmidrule(ll){3-8}%
                \cmidrule(ll){9-14}%
            },
            after row={
                \cmidrule(ll){2-14}%
                {\bfseries Ensemble}%
                &%
                {%
                    \begin{tabular}{c}
                        \texttt{max}%
                        \
                        \texttt{depth}%
                    \end{tabular}
                }%
                \\
                \midrule
            }
        },
        every last row/.style={%
            after row=\bottomrule
        },
        every row no 1/.style={%
            before row=\cmidrule(ll){1-14}
        },
        every nth row={2}{
            after row=\cmidrule(ll){1-14}
        },
        multicolumn names,
    ]{\loadedtable}
}

\NewDocumentCommand{\showLone}{m m m}{%
    \pgfplotstableread[
        col sep=comma,
        header=true,
        comment chars={\#},
        columns/dataset/.style={string type}
    ]{#1}\loadedtable 

    \pgfplotsinvokeforeach{#2,#3} {
        \pgfplotstableset{
            columns/n-active-estimators-##1/.style={n-active-estimators},
            columns/fidelity-##1/.style={fidelity},
            columns/elapsed-time-##1/.style={elapsed-time},
            columns/accuracy-##1/.style={accuracy},
            columns/n-oracle-calls-##1/.style={n-oracle-calls},
        }
    }

    \xdef\thiscolumns{dataset,}
    \foreach \M [count=\i] in {#2,#3} {
        \xdef\thiscolumns{\thiscolumns n-active-estimators-\M,}
        \xdef\thiscolumns{\thiscolumns n-oracle-calls-\M,}
        \xdef\thiscolumns{\thiscolumns fidelity-\M,}
        \xdef\thiscolumns{\thiscolumns accuracy-\M,}
        \xdef\thiscolumns{\thiscolumns elapsed-time-\M}
        \ifnum\i<2
        \xdef\thiscolumns{\thiscolumns,}
        \fi
    }
    
    \pgfplotstabletypeset[
        columns/.expand once=\thiscolumns,
        columns/dataset/.style={dataset},
        every head row/.style={
            before row={
                \toprule
                & \multicolumn{5}{c}{\(M=#2\)}%
                & \multicolumn{5}{c}{\(M=#3\)}%
                \\
                \cmidrule(lr){2-6}
                \cmidrule(lr){7-11}
            },
            after row=\midrule
        },
        every last row/.style={after row=\bottomrule},
        multicolumn names,
        sort,
        sort cmp={string <},
        sort key=dataset,
    ]{\loadedtable}
}

\NewDocumentCommand{\showLoneAll}{m m m m}{
    \pgfplotstableread[
        col sep=comma,
        header=true,
        comment chars={\#},
        columns/ensemble/.style={string type}
        columns/dataset/.style={string type}
    ]{#1}\loadedtable 

    \pgfplotsinvokeforeach{#2,#3} {
        \pgfplotstableset{
            columns/n-active-estimators-##1/.style={n-active-estimators},
            columns/fidelity-##1/.style={fidelity},
            columns/elapsed-time-##1/.style={elapsed-time},
            columns/accuracy-##1/.style={accuracy},
            columns/n-oracle-calls-##1/.style={n-oracle-calls},
        }
    }

    \xdef\thiscolumns{ensemble, max-depth, dataset}
    \foreach \M [count=\i] in {#2,#3} {
        \xdef\thiscolumns{\thiscolumns, n-active-estimators-\M}
        \xdef\thiscolumns{\thiscolumns, n-oracle-calls-\M}
        \xdef\thiscolumns{\thiscolumns, fidelity-\M}
        \xdef\thiscolumns{\thiscolumns, accuracy-\M}
        \xdef\thiscolumns{\thiscolumns, elapsed-time-\M}
    }
    
    \pgfplotstabletypeset[
        columns/.expand once=\thiscolumns,
        columns/dataset/.style={dataset},
        columns/ensemble/.style={
            ensemble,
            assign cell content/.code={%
                \pgfmathtruncatemacro\row{\pgfplotstablerow}%
                \pgfmathparse{int(mod(\row,22))}%
                \ifnum\pgfmathresult=0%
                \pgfkeyssetvalue{/pgfplots/table/@cell content}{%
                    \multirow{22}{*}{####1}
                }%
                \else
                \pgfkeyssetvalue{/pgfplots/table/@cell content}{}%
                \fi%
            },
        },
        columns/max-depth/.style={
            column name = {%
                \texttt{max}%
                \ %
                \texttt{depth}%
            },
            column type = {c},
            fixed,
            fixed zerofill,
            precision=0,
            assign cell content/.code={%
                \pgfmathparse{int(mod(\pgfplotstablerow,11))}%
                \ifnum\pgfmathresult=0%
                \pgfkeyssetvalue{/pgfplots/table/@cell content}{%
                    \multirow{11}{*}{####1}
                }%
                \else
                \pgfkeyssetvalue{/pgfplots/table/@cell content}{}%
                \fi
            },
        },
        every head row/.style={
            before row={
                #4\\
                \toprule
                &%
                &%
                & \multicolumn{5}{c}{\(M=#2\)}%
                & \multicolumn{5}{c}{\(M=#3\)}%
                \\
                \cmidrule(ll){4-8}
                \cmidrule(ll){9-13}
                \endfirsthead
                \toprule
                &%
                &%
                & \multicolumn{5}{c}{\(M=#2\)}%
                & \multicolumn{5}{c}{\(M=#3\)}%
                \\
                \cmidrule(ll){4-8}
                \cmidrule(ll){9-13}
                \model{Ensemble}%
                & \texttt{max} \texttt{depth}%
                & \multicolumn{1}{c}{%
                    \textbf{Dataset}%
                }%
                & \(\left\| w \right\|_{0}\)%
                & \(\noracle\)%
                & \multicolumn{1}{c}{%
                    \textbf{\faithmetric}%
                }%
                & \multicolumn{1}{c}{%
                    \textbf{\accuracy}%
                }%
                & \textbf{Time} (\SI{}{\second})%
                & \(\left\| w \right\|_{0}\)%
                & \(\noracle\)%
                & \multicolumn{1}{c}{%
                    \textbf{\faithmetric}%
                }%
                & \multicolumn{1}{c}{%
                    \textbf{\accuracy}%
                }%
                & \textbf{Time} (\SI{}{\second})%
                \\
                \midrule
                \endhead%
            },
            after row=\midrule
        },
        every last row/.style={
            after row=\bottomrule
        },
        every nth row={11}{%
            before row=\cmidrule(ll){2-13}%
        },
        every nth row={22}{%
            before row=\cmidrule(ll){1-13}%
        },
        every nth row={\nLinesPerLongTablePage}{
            before row={
                \cmidrule(ll){1-13}
                \pagebreak
            }
        },
        multicolumn names,
        begin table=\begin{longtable},
        end table=\end{longtable}
    ]{\loadedtable}
}

\pgfkeys{
    /pgf/number format/.cd,
    1000 sep={\,}
}

%% file: sections/0-abstract.tex
\begin{abstract}
    Tree ensembles, including boosting methods, are highly effective and widely used for tabular data. However, large ensembles lack interpretability and require longer inference times. We introduce a method to prune a tree ensemble into a reduced version that is ``functionally identical'' to the original model. In other words, our method guarantees that the prediction function stays unchanged for any possible input. As a consequence, this pruning algorithm is lossless for any aggregated metric. We formalize the problem of functionally identical pruning on ensembles, introduce an exact optimization model, and provide a fast yet highly effective method to prune large ensembles. Our algorithm iteratively prunes considering a finite set of points, which is incrementally augmented using an adversarial model. In multiple computational experiments, we show that our approach is a ``free lunch'', significantly reducing the ensemble size without altering the model's behavior. Thus, we can preserve state-of-the-art performance at a fraction of the original model's size.
\end{abstract}

%% file: sections/1-introduction.tex
\section{Introduction}

Ensembles remain one of the most effective and commonly utilized machine learning models. Among them, tree ensembles such as random forests and boosting are known to achieve state-of-the-art performance on tabular data \citep{Grinsztajn2022, McElfresh2024neural}. They also offer greater interpretability compared to large neural networks. Theory and practice indicate that superior performance is achieved with large ensembles, i.e., forests with many trees \citep{Biau2016,Probst2017,Buschjager2021improving}. However, large ensembles lead to large memory requirements and inference times, which can quickly become a bottleneck when embedding models into hardware such as microcontrollers or smartphones. Further, large ensemble models lack interpretability due to the complex interaction of their components.

Ensemble pruning denotes the process of reducing the size of a trained ensemble model. This can be understood as removing nodes from the trees or removing trees entirely from the forest. Early approaches studied how to prune random forests made of shallow trees, i.e., trees that were themselves pruned beforehand \citep[see, e.g.,][]{Margineantu1997pruning, Martinez2006pruning}. These works showed that pruning could improve the test error and out-of-sample generalization. However, this is not true with boosted ensembles, since the trees are typically already shallow. Hence, most recent studies seek a trade-off between the amount of pruning and the impact on test error \citep{Liu2023forestprune, Amee2023novel}.
This trade-off may be profitable in some situations but there is no guarantee that the model will perform well on new data.

\begin{figure}[ht!]
    \centering
    \makeatletter%
    \if@twocolumn%
        \setlength{\figurewidthx}{0.9\linewidth}%
    \fi%
    \makeatother%
    \includegraphics[width=\figurewidthx]{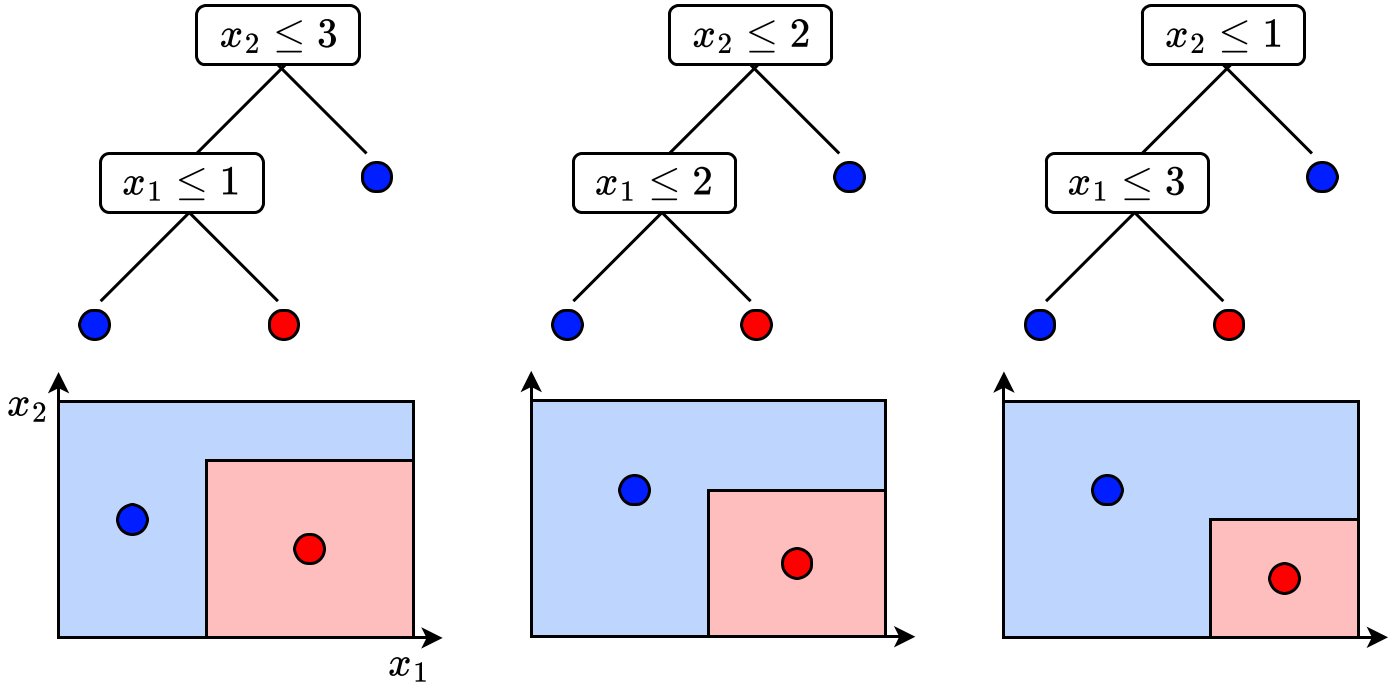}
    \caption{A small ensemble made of three trees with equal weights. This ensemble can be pruned without any change in its prediction function by removing the first and third trees.}%
    \label{fig:prune-example}%
\end{figure}

In this paper, we completely avoid this trade-off by pruning ensemble models \emph{without any change in their behavior}. We call this a ``functionally-identical'' pruning since the prediction functions of the pruned ensemble and the original one are identical. This has two main advantages. First, as demonstrated in this paper, the resulting optimization problems can be very efficiently solved. Second, such pruned models give a ``free lunch'': they achieve the exact same performance as the original model at a fraction of its size. This provides multiple advantages regarding memory, inference time, and interpretability. An example ensemble that can be pruned without any change in its prediction function is shown in \cref{fig:prune-example}.

This paper introduces the algorithm \model{FIPE}, standing for \textbf{F}unctionally \textbf{I}dentical \textbf{P}runing of \textbf{E}nsembles. \model{FIPE} iterates between a \textit{pruning model} and a \textit{separation oracle}. The pruning model identifies candidate pruned ensembles on a finite set of points. The separation oracle checks whether the pruned model is functionally identical to the original model on the entire feature space. If this is not the case, new points are added to the pruning set. Our iterative algorithm is such that the final pruned ensemble is certifiably equivalent to the original one on the entire feature space.

We make the following contributions:
\setlist{nolistsep}
\begin{enumerate}[
    label=(\roman*),
    labelindent=0pt,
    labelwidth=\widthof{\ref{last-item}},
    leftmargin=!
]
    \setlength{\itemsep}{1pt}

    \item We formalize the problem of functionally identical pruning of tree ensembles and characterize its computational complexity.

    \item We present \model{FIPE}, our iterative algorithm and its two main components: the pruner, and the oracle. We prove that the algorithm terminates in a finite number of steps.

    \item We investigate two variants of the pruning model: an exact combinatorial optimization model based on the \(\left\|\cdot\right\|_{0}\) norm that guarantees that the pruned model is of minimal size, and a fast approximate algorithm based on the \(\left\|\cdot\right\|_{1}\) norm that is not necessary of minimal size but scales to large datasets. We further provide an efficient pruning oracle inspired by recent works on counterfactual explanations.

    \item We apply \model{FIPE} on boosted ensembles \model{AdaBoost}, \model{LightGBM} and \model{XGBoost} as well as random forests. Our extensive experiments demonstrate that significantly smaller ensembles can be identified across various real-world datasets. A central finding of this study is identifying that many base learners of boosted ensembles are superfluous. Consequently, an integrated pruning and reweighting approach can substantially reduce their size while maintaining their predictive performance. \label{last-item}
\end{enumerate}

%% file: sections/2-problem.tex
\section{Problem Statement}%
\label{sec:prob}%

Let \(\left\{\left(x_i, y_i\right)\right\}_{i=1}^n\) be a labeled set of observations, where \(x \in \mathcal{X} \subseteq \mathbb{R}^p\) is a feature vector. We focus on classification problems with \(C\) classes. Denote by \(\llbracket a, b \rrbracket\) the set of integers between \(a\) and \(b\). Any class \(y\) belongs to \(\left\llbracket 1, C\right\rrbracket\). 

\subsection{Classification ensembles}%

A classification ensemble is a weighted set of classifiers \(\smash{\left\{\left(h_m, \alpha_m\right) \right\}_{m=1}^M}\) where each classifier \(h_m: \mathcal{X} \rightarrow [0, 1]^{C}\) provides a score vector for any input \(x\), and \(\alpha_m\) is the weight of classifier \(m\). The ensemble prediction function is a map \(H: \mathcal{X} \times \mathbb{R}_{\ge 0}^M \rightarrow \left\llbracket 1, C\right\rrbracket\) that assigns a class to any input \(x\) using a voting scheme, expressed generally as:
\begin{equation}%
    \label{eq:majority-voting}%
    H(x; \alpha) = \argmax_{\class \in \left\llbracket 1, C\right\rrbracket} \sum\nolimits_{m=1}^M \alpha_m h_m^{(\class)}(x),
\end{equation}
where \(\smash{h_m^{(\class)}(x)}\) is the score predicted by classifier of index~\(m\) for class \(\class\). Any deterministic tie-breaking rule can be used to break the ties in Equation~\eqref{eq:majority-voting} if the \(\argmax\) is set-valued.

The computation of the score of a tree depends on the nature of the ensemble. For instance, AdaBoost \citep{Hastie2009multi} and the original random forest algorithm of \citet{Breiman2001} use the so-called hard-voting criterion in which each classifier prediction \(\smash{h_m^{(\class)}(x) \in \{0, 1\}}\) is binary.

\subsection{Functionally-identical pruning}%
\label{sec:definitions}%
We study how to prune classification ensembles while guaranteeing that the pruned model is functionally identical to the original model. This is formalized in the following definition.
\begin{definition}%
    \label{def:faithful}%
    A pruned ensemble with weights \(w\) is \emph{functionally identical} to the original ensemble on the entire feature space \(\mathcal{X}\) if \(H(x; w) = H(x; \alpha) \) for all \(x \in \mathcal{X}\).
\end{definition}

The pruned ensemble with minimum size is obtained by solving:
\begin{equation}%
    \label{opt:fipe}%
    \min_{w \ge 0}\; \left\|w\right\|_{0}%
    \; \colon%
    \; H(x; w) = H(x; \alpha),%
    \; \forall x \in \mathcal{X}.%
\end{equation}

Since any learner with a zero weight is effectively pruned, Problem~\eqref{opt:fipe} minimizes the number of active learners. The minimizer of Problem~\eqref{opt:fipe} is not only certifiably functionally identical, but it is also of minimal size. That is, it is guaranteed to be the smallest reweighted ensemble that is functionally identical to the original one.

The constraint in Problem~\eqref{opt:fipe} ensures that the pruned model is functionally identical to the original model on the entire space. Functionally-identical pruning is also known as ``faithful'' pruning in \citet{Vidal2020born}. \onlyextended[W]{Hence, w}e use equivalently both words in the rest of the paper.
\begin{proposition}\label{prop:complexity}
    For additive tree ensembles, i.e., a broad class including random forests and boosting methods, Problem~\eqref{opt:fipe} is NP-hard.
\end{proposition}
The proof is given in \onlyextended[Appendix A (see extended version)]{\cref{app:method}}. \cref{prop:complexity} states that optimal faithful pruning is a difficult task in general. Nevertheless, this paper shows that Problem~\eqref{opt:fipe} can be solved efficiently for large ensembles on real-world datasets, and provides effective heuristics otherwise. Our algorithms are presented in the next section.

%% file: sections/3-prune.tex
\section{Pruning Algorithms}%
\label{sec:entire}%

To solve the faithful pruning problem, \model{FIPE} iterates between solving a \emph{pruning problem} on a finite set of points and a \emph{separation oracle}. This is illustrated in~\cref{fig:fipe-iterative}.
\begin{figure}[ht]%
    \centering
    \input{figures/fipe.tex}%
    \caption{\model{FIPE} iterates between the pruning model and the separation oracle until it returns the set of weights of the pruned model.}%
    \label{fig:fipe-iterative}%
\end{figure}

\model{FIPE} is thus made of two essential components: (i)~a faithful pruner \(\model{Pruner}\) that returns the set of weights \(\smash{{\left\{w_m\right\}}_{m=1}^M}\) of the pruned ensemble faithful to the original ensemble on a given finite set of points \(\finiteset\subset\mathcal{X}\), and (ii)~a separation oracle \(\model{Oracle}\) that returns a set of point \(\newpoints \subset\mathcal{X} \) for which the pruned ensemble with weights \(w\) and the original ensemble with weights \(\alpha\) differ. If the oracle cannot find a separating point, the two ensembles are functionally identical over the entire feature space. The algorithm then returns the set of weights of the pruned ensemble. This algorithm is presented in \cref{alg:fipe}.

\begin{algorithm}[tb]%
    \caption{Faithful pruning algorithm}%
    \label{alg:fipe}%
    \input{algorithms/fipe}
\end{algorithm}%

\begin{theorem}%
    \label{thm:fipe-terminates}%
    For tree ensembles, \model{FIPE} terminates after a finite number of calls to the \model{Oracle}.%
\end{theorem}%

The proof is given in \onlyextended[Appendix A (see extended version)]{\cref{app:method}}.

\subsection{Pruning on a finite set of points}%
\label{subsec:finite-pruning}%
Let  \(\finiteset = \smash{{\left\{\left(x_i, y_i\right)\right\}}_{i=1}^N}\) a finite set of points \(\finiteset\subset\mathcal{X}\). The pruning model returns the set of weights of the pruned ensemble faithful to the original ensemble on \(\finiteset\). 

Denote by \( \class_i = H \left(x_i, \alpha\right) \) the class of the point \(x_i\) given by the original ensemble with weights \(\alpha\). Note that the classes \(\class_i\) are not necessarily equal to the original classes \(y_i\) depending on the training of the ensemble \(H\). Pruning on a finite set of points is a special case of Problem~\eqref{opt:fipe} when $\mathcal{X}$ is finite. The constraint of Problem~\eqref{opt:fipe} can be reformulated to ensure that the pruned ensemble is faithful to the original ensemble on \(\finiteset\). It can be written as:
\begin{equation}%
    \label{eq:finite-pruning-problem-cons}%
    \sum\nolimits_{m=1}^M w_m\left(h^{(\class_i)}_m\left(x_i\right)-h^{(\class)}_m\left(x_i\right)\right)>0.%
\end{equation}%
It ensures that the pruned ensemble gives a higher weighted score for the class \(\class_i\) than for the class \(\class\) for each point \(x_i\) and each class \(\class \neq \class_i\). 

Equation~\eqref{eq:finite-pruning-problem-cons} is difficult to handle numerically since it involves a strict inequality. However, using the same idea as in a support vector machine, we express it equivalently as:
\begin{equation}%
    \label{eq:finite-pruning-problem-cons-linear}%
    \sum\nolimits_{m=1}^M w_m\left(h^{(\class_i)}_m\left(x_i\right)-h^{(\class)}_m\left(x_i\right)\right)\ge1.
\end{equation}%

\paragraph{Minimal-size faithful pruner.} The faithful pruning problem on a finite set of points is a combinatorial optimization problem given by:
\begin{subequations}%
    \label{eq:finite-pruning-problem-integer}%
    \begin{align}%
        &\min_{w \ge 0, \, u \in \left\{ 0, 1 \right\}^M}\;
            \sum\nolimits_{m=1}^M u_m
        \\
        &w_m \leq W u_m,\;
        \forall m \in \left\llbracket 1, M \right\rrbracket,%
        \\
        &\sum\nolimits_{m=1}^M w_m\left(h^{(\class_i)}_m\left(x_i\right)-h^{(\class)}_m\left(x_i\right)\right)\ge1,%
        \label{eq:finite-pruning-problem-integer-cons}%
        \\
        &\forall i \in \left\llbracket 1, N \right\rrbracket,\,%
            \forall \class\in\left\llbracket 1, C \right\rrbracket,\,%
            \class \neq \class_i.
        \nonumber
    \end{align}%
\end{subequations}%
This problem minimizes the \(\left\|\cdot\right\|_{0}\) norm of the weights. It uses the binary variables~\(u_m\) to indicate whether the estimator \(m\) is active or not, and a parameter \(W\), which is an upper bound on the weights of the active estimators. A minimizer of Problem~\eqref{eq:finite-pruning-problem-integer} is both faithful to the original ensemble on \(\finiteset\) and of certifiably minimal size according to the \(\left\|\cdot\right\|_{0}\) norm.

\paragraph{Efficient approximation.} While Problem~\eqref{eq:finite-pruning-problem-integer} ensures finding the sparsest model, it might be expensive to solve numerically since its objective function with \(\left\|\cdot\right\|_{0}\) is hard to linearize. Hence, we propose an efficient approximation by replacing the \(\left\|\cdot\right\|_{0}\) norm with the \(\left\|\cdot\right\|_{1}\) norm. This problem is a linear programming problem given by:
\begin{subequations}%
    \label{eq:finite-pruning-problem-relaxation}%
    \begin{align}%
        \mathrlap{\min_{w \ge 0} \; \left\|w\right\|_{1} = \sum\nolimits_{m=1}^M w_m} \\
        & \sum\nolimits_{m=1}^M w_m\left(h^{(\class_i)}_m\left(x_i\right)-h^{(\class)}_m\left(x_i\right)\right)\ge1,%
        \label{eq:finite-pruning-problem-relaxation-cons} \\
        & \forall i \in \left\llbracket 1, N \right\rrbracket,\, \forall \class\in\left\llbracket 1, C \right\rrbracket,\, \class \neq \class_i.\nonumber
    \end{align}%
\end{subequations}%
Problem~\eqref{eq:finite-pruning-problem-relaxation} can be solved very efficiently using standard linear programming solvers. The faithfulness constraint of the problem is unchanged so that any solution of Problem~\eqref{eq:finite-pruning-problem-relaxation} is guaranteed to remain faithful to the original ensemble on~\(\finiteset\).

%% file: figures/fipe.tex
\begin{tikzpicture}[
      node distance=2cm,
      auto,
      >=latex,
      thick,
]
\tikzstyle{bloc} = [
      rectangle,
      draw,
      fill=blue!10,
      text width=4em,
      text centered,
      rounded corners,
      minimum height=2em
]
\tikzstyle{oracle} = [
      rectangle,
      draw,
      fill=red!10,
      text width=4em,
      text centered,
      rounded corners,
      minimum height=2em
]
\tikzstyle{union} = [
      rectangle,
      draw,
      fill=green!10,
      text centered,
      rounded corners,
      minimum height=1em
]
\node (set) {\(\mathcal{D}^0\)};
\node (pruner) [
      anchor=west,
      bloc,
      right of=set
] {\model{Pruner}};
\node (oracle) [
      oracle,
      right=2cm of pruner
] {\model{Oracle}};
\node (stop) [
      right of=oracle
] {\(w^k\)};
\draw[->] (set)
      --
      (pruner);
\draw[->] (pruner)
      -- node [above] (wk) {\(w^k\)}
      (oracle);
\node (union) [
      union,
      below = 1cm of wk
] {\(\mathcal{D}^{k+1} = \mathcal{D}^k \cup \newpoints^k \)};
\draw[->] (oracle)
      |- node [right] {\(\newpoints^k\)}
      (union);
\draw[->] (union)
      -| node [left] {\(\mathcal{D}^{k+1}\)}
      (pruner);
\draw[->] (oracle)
      -- node [above] {\(\varnothing\)}
      (stop);
\end{tikzpicture}

%% file: algorithms/fipe.tex
\begin{algorithmic}[1]
\STATE\(\smash{\mathcal{D}^0\gets\text{An arbitrary finite set of points of }\mathcal{X}}\)
\STATE\(\smash{w^0\gets\model{Pruner}\left(\mathcal{D}^0\right)}\)
\STATE\(k\gets0\)
\WHILE{\(\smash{\model{Oracle}\left(w^k\right)}\) is not \(\varnothing\)}
    \STATE\(\smash{\newpoints^k \gets \model{Oracle}\left(w^k\right)}\)
    \STATE\(\smash{\mathcal{D}^{k+1} \gets \mathcal{D}^k\cup \newpoints^k}\)
    \STATE\(\smash{w^{k+1}\gets\model{Pruner}\left(\mathcal{D}^{k+1}\right)}\)
    \STATE\(k\gets k+1\)
\ENDWHILE%
\RETURN\(w^k\)
\end{algorithmic}

%% file: sections/4-oracle.tex
\subsection{Separation oracle for tree ensembles}%
\label{subsec:separation-oracle}%

This section presents the separation oracle used in \model{FIPE}. The goal of the oracle is twofold: (i)~it certifies that a pruned model is functionally equivalent to the original model on the entire feature space, or (ii)~it provides a set of points for which the two models differ. These two goals can be achieved jointly by formulating an optimization model that maximizes the misclassification between the pruned and original model over the feature space.

Let \(\class\) and \(y\) be two fixed classes. The separation problem over these two classes can be formulated as:
\begin{equation}%
    \label{eq:sep-pair}%
    \max_{x\in\mathcal{X}}\,\smashoperator{\sum_{m=1}^M} w_m\left(h^{(\class)}_m\left(x\right)-h^{(y)}_m\left(x\right)\right)\,%
    \colon \, y = H\left(x, \alpha\right).%
\end{equation}%

This problem maximizes the difference between the score of class \(\class\) and class \(y\) according to the pruned ensemble. The constraint of Problem~\eqref{eq:sep-pair} ensures that the original ensemble predicts class \(y\) for the point \(x\). If the value of Problem~\eqref{eq:sep-pair} is positive, the pruned ensemble predicts class \(\class\) whereas the original ensemble predicts  \(y\). Hence, a separating point is such that it has a positive value according to Problem~\eqref{eq:sep-pair}.
\begin{remark}
    Since Problem~\eqref{eq:sep-pair} maximizes the difference in prediction between two ensembles, it can be interpreted as a variation of the adversarial example or counterfactual explanation problem.
\end{remark}

If the objective function is non-positive across all points in the feature space, both models are certified to be equivalent on the entire feature space. In practice, we solve the separation problem for each pair of classes \(\class\) and \(y\). Depending on the algorithm used to solve Problem~\eqref{eq:sep-pair}, several points may be found that have a positive score difference. In \model{FIPE}, we add all points with a positive score difference \(\newpoints\). When \(\newpoints\) is empty, the algorithm terminates.

The complexity of Problem~\eqref{eq:sep-pair} lies in representing the classification function of the trees with linear variables and constraints. To ensure that the original ensemble predicts class \(y\) for point \(x\), we can replace the constraint of Problem~\eqref{eq:sep-pair} by the following inequality:
\begin{equation}%
    \label{eq:sep-pair-cons-inequality}%
    \sum\nolimits_{m=1}^M \alpha_m \left(h^{(y)}_m\left(x\right)- h^{(y')}_m\left(x\right)\right) > 0, \,\forall y'\neq y.%
\end{equation}%

These constraints ensure that the original ensemble gives a higher score for class \(y\) than for any other class \(y' \neq y\), making \(y\) the predicted class for point \(x\) by the original ensemble. Again, this constraint is ill defined because of the strict inequality. By introducing a sufficiently small \(\varepsilon > 0\), the separation problem can be equivalently formulated as:
\begin{subequations}%
    \label{eq:sep-pair-diff}%
    \begin{alignat}{3}%
        \mathrlap{\max_{x\in\mathcal{X}}\; \sum\nolimits_{m=1}^M w_m\left(h^{(\class)}_m\left(x\right)-h^{(y)}_m\left(x\right)\right)}%
        \label{eq:sep-pair-diff-obj} \\
        & \sum\nolimits_{m=1}^M \alpha_m \left(h^{(y)}_m\left(x\right)- h^{(y')}_m\left(x\right)\right)& \ge\varepsilon,\;
        &&\forall y'\neq y.%
        \label{eq:sep-pair-diff-cons}
    \end{alignat}%
\end{subequations}%

\paragraph{Separation for tree ensembles.} Problem~\eqref{eq:sep-pair-diff} is nonlinear in general. However, for additive tree ensembles, we can linearize it efficiently using a procedure similar to the one used in \citet{Parmentier2021}. This consists in introducing a set of variables and constraints to track the path of the separating point \(x\) in all trees of the ensembles.

For each tree \(m\), let \(\nodes_{m}\) be the set of its internal nodes, \(\leaves_{m}\) the set of its leaves, and~\(d_m\) its maximum depth. For each node \(\node\in\nodes_{m}\), let \(\treeleft(\node)\) and \(\treeright(\node)\) be its left and right children respectively, and let \(\treedepth(\node)\) be its depth in the tree \(m\). Let \(\treeroot(m)\) denote the root of tree \(m\). Let \(\flowvar_{m, \node}\) be a binary variable indicating if node \(\node\in\nodes_{m}\cup \leaves_{m}\) is in the path of the point \(x\) in the tree \(m\).

To linearize the tree prediction functions, we introduce a binary variable \(\lambda_{m, d}\) to indicate whether point \(x\) goes to the left child at depth \(d\) of the tree \(m\). The path consistency constraints in the tree \(m\) can then be expressed as:
\begin{subequations}%
    \begin{alignat}{3}%
        \flowvar_{m, \treeroot(m)} &= 1,\,\label{eq:sep-pair-cons-lin-root} \\
        \flowvar_{m, \treeleft(\node)} + \flowvar_{m, \treeright(\node)} &= \flowvar_{m, \node},\,%
        && \forall \node\in\nodes_{m},%
        \label{eq:sep-pair-cons-lin-children} \\
        \smashoperator[r]{%
        \sum_{\mathclap{\quad\qquad\qquad\node\in\nodes_{m}:\, \treedepth(\node) = d}}
        }\quad\flowvar_{m, \treeleft(\node)} &= \lambda_{m, d},\,%
        &&\forall d\in\left\llbracket 0, d_m \right\rrbracket.%
        \label{eq:sep-pair-cons-lin-final}%
    \end{alignat}%
\end{subequations}%

Constraint \eqref{eq:sep-pair-cons-lin-root} ensures that the root of tree \(m\) is in the path of point \(x\). Constraint \eqref{eq:sep-pair-cons-lin-children} ensures that one of the children of node \(\node\) is in the path of point \(x\) if node \(\node\) is in the path. It also ensures that the path of point \(x\) in tree \(m\) at depth \(d\) goes to the left child if \(\lambda_{m, d} = 1\). Using these variables, we can formulate Problem~\eqref{eq:sep-pair-diff} equivalently as:
\begin{alignat}{2}%
    \mathllap{%
    \max_{\flowvar \ge 0}\;%
    \smashoperator[r]{%
    \sum_{m=1}^M}%
    \sum_{\node\in\leaves_{m}}%
    w_m \left(h^{(c)}_m\left(x\right)- h^{(y)}_m\left(x\right)\right) \flowvar_{m, \node}}%
    \label{eq:sep-pair-cons-lin}\\
    \smashoperator[r]{%
    \sum_{m=1}^M}\;%
    \sum_{\node\in\leaves_{m}}%
    \alpha_m \left(h^{(y)}_m\left(x\right)- h^{(y')}_m\left(x\right)\right) \flowvar_{m, \node}\ge \varepsilon,\,%
    &\forall y'\neq y. \nonumber
\end{alignat}%

\begin{remark}
    \label{rem:cell}
    Problem~\eqref{eq:sep-pair-cons-lin} is not optimizing over the variable \(x\) anymore, but only over the variables that track the path over the trees. Indeed, for tree ensembles, the prediction function is piecewise-constant. Hence, it is sufficient to identify the set of leaves to separate the pruned and original ensembles. A separating point is directly determined by taking the center of the cell defined by the leaf variables.
\end{remark}
Further, observe that the binary variables \(\flowvar_{m, \node}\) can be defined as continuous in \([0, 1]\), as they will be forced to binary values due to Constraints \eqref{eq:sep-pair-cons-lin-root}--\eqref{eq:sep-pair-cons-lin-final}.

\paragraph{Feature consistency.} To ensure the value of \(x\) is consistent across the trees, we need to add feature consistency variables and constraints. Each feature type has to be handled separately. We explain below how to ensure the consistency of continuous features, and in \onlyextended[Appendix A (see extended version)]{\cref{app:method}} how to handle categorical and binary features.

Let \(j\) be a continuous feature and let \(\left\{t_{j, 1}, t_{j, 2}, \ldots, t_{j, R_j}\right\}\) be the set of thresholds that the nodes of the original ensemble are splitting on. We consider without loss of generality that this set is sorted. For each continuous feature \(j\) and \(r\in \left\llbracket 1, R_j \right\rrbracket\), we define the continuous auxiliary variables \(\mu_{j, r} \in [0, 1]\). These variables indicate whether the new point is on the left or right-side of a level. They are constrained such that \(\mu_{j, r} = 0 \Leftrightarrow x_j \in \left]-\infty,t_{j, r}\right]\) by imposing:
\begin{alignat}{2}%
    \mu_{j, r} &\ge \mu_{j, r+1},\, && \forall r\in\left\llbracket 1, R_j-1\right\rrbracket \label{eq:sep-pair-cons-lin-continuous-order}%
\end{alignat}%

For each tree \(m\), let \(\nodes_{m}^{j, r}\) be the set of nodes that split on feature \(j\) at threshold \(t_{j, r}\) for \(r\in \left\llbracket 1, R_j \right\rrbracket\). The following constraints ensure that the value of \(\mu_{j, r}\) is consistent across the nodes of tree \(m\):
\begin{subequations}%
    \begin{alignat}{3}%
        &\flowvar_{m, \treeleft(\node)} &&\le 1-\mu_{j, r},\; && \forall r \in \left\llbracket 1, R_j \right\rrbracket, \forall \node \in \nodes_{m}^{j, r}, \label{eq:sep-pair-cons-lin-continuous-left} \\
        &\flowvar_{m, \treeright(\node)} &&\le \mu_{j, r},\; && \forall r \in \left\llbracket 1, R_j \right\rrbracket, \forall \node \in \nodes_{m}^{j, r}. \label{eq:sep-pair-cons-lin-continuous-right}%
    \end{alignat}%
\end{subequations}%

Constraint \eqref{eq:sep-pair-cons-lin-continuous-left} ensures that if the value of \(x_j\) is in the interval \(\left]t_{j, r}, \infty\right[\), then at node \(\node\) we cannot go to the left child. Similarly, constraint \eqref{eq:sep-pair-cons-lin-continuous-right} ensures that if the value of \(x_j\) is in the interval \(\left]-\infty, t_{j, r}\right]\), then at node \(\node\) we can not go to the right child.

Our separation oracle resembles closely the problem of finding counterfactual explanations \citep{Parmentier2021} or classifier evasions \citep{Kantchelian2016evasion}. Yet, optimizing over the set of leaves (see~\cref{rem:cell}) instead of the continuous variable \(x\) offers numerical benefits. Determining the exact value of $x$, as needed in counterfactual explanation, requires checking for strict inequalities when going right on a threshold.

%% file: sections/5-numerical.tex
\section{Computational Experiments}%
\label{sec:experiments}%

\begin{table*}[t]
    \centering
    \small
    \input{tables/ab/ab-l0-vs-l1}
    \caption{Comparison of \model{FIPE}\(-\left\| \cdot \right\|_{0}\) and \model{FIPE}\(-\left\| \cdot \right\|_{1}\) for pruning \model{AdaBoost} ensembles with \(\pgfmathprintnumber{\LZEROEST}\) learners. The table shows the number of learners in the pruned ensemble \(\left\| w \right\|_{0}\), test faithfulness to the original ensemble \(\faithmetric\), test accuracy \(\accuracy\), and rounded number of oracle calls \(\noracle\). All results are averaged over five repetitions with different train/test splits.}
    \label{tab:ab-l0-vs-l1}
\end{table*}

We now study the value of \model{FIPE} for pruning large training ensembles while maintaining faithfulness. We investigate how much \model{FIPE} can prune ensembles and its scalability over several datasets. We analyze the behavior of \model{FIPE} in terms of what trees are removed, and how pruning is impacted by the size of the original ensemble. Finally, we compare our approach to baselines from the recent literature.

In this section, we focus our experiments on \model{AdaBoost} classifiers. Experiments with other boosted ensembles, such as \model{XGBoost} and \model{LightGBM}, are provided in \onlyextended[Appendix B (see extended version)]{\cref{app:experiments}} with essentially the same conclusions. We also study random forests in this appendix. We perform our analyses across \(\pgfmathprintnumber{\NDATASETS}\) datasets commonly used in previous studies. In each experiment, we split the data set into a training dataset (\pgfmathprintnumber{\TRAINSPLIT}\%) and a test dataset (\pgfmathprintnumber{\TESTSPLIT}\%) to measure faithfulness and test accuracy. This process is repeated over five different random seeds. The characteristics of the datasets are presented in \cref{tab:datasets}: their number of samples~\(n\), number of features~\(p\) including the number of numerical~\(p_N\) and binary~\(p_B\) features, and number of classes~\(C\).

\begin{table}[htb]
    \centering
    \small
    \input{tables/datasets}
    \caption{Characteristics of the data sets}
    \label{tab:datasets}
\end{table}

All experiments are implemented in \package{python}. We used \package{scikit-learn} for training the ensembles. All optimization problems are solved to global optimality using the commercial solver \package{Gurobi v11.0}. The experiments are conducted on a computing grid. Each experiment utilizes a single core of an Intel Xeon Gold 6258R CPU running at \SI{2.7}{\giga\hertz} and is provided with \SI{4}{\giga\byte}~RAM. \onlyextended{\model{FIPE} is provided as a standalone \package{python} package at \fiperepo. To reproduce the results of this paper, we provide the code and the datasets at \experimentsrepo.}

\subsection{Main results}%
\label{subsec:main-results}%
First, we evaluate how much \model{FIPE} is able to prune tree ensembles while certifying faithfulness to the original model. We compare both the certifiable minimal-size and fast but approximate, respectively denoted by \model{FIPE}\(-\left\| \cdot \right\|_{0}\) and \model{FIPE}\(-\left\| \cdot \right\|_{1}\). We measure the number of learners in the pruned ensemble, the faithfulness to the original ensemble on the test set denoted \(\faithmetric\), the test accuracy \(\accuracy\), and the number of oracle calls rounded to its upper integer denoted by \(\noracle\). All performance metrics are average over the five repetitions.

The results are presented in \cref{tab:ab-l0-vs-l1} for ensembles of \(\pgfmathprintnumber{\LZEROEST}\) learners. They show that \model{FIPE} is consistently able to prune \model{AdaBoost} ensembles while maintaining faithfulness. Further, we observe that \model{FIPE}\(-\left\| \cdot \right\|_{1}\) achieves the same pruned size as \model{FIPE}\(-\left\| \cdot \right\|_{0}\) for all but one dataset, while having significantly faster pruning times.

\begin{result}
    Using \(\left\| \cdot \right\|_{1}\) in \model{FIPE} typically yields pruning performance similar to \(\left\| \cdot \right\|_{0}\) but with significantly faster runtimes.
\end{result}

Because \model{FIPE}\(-\left\| \cdot \right\|_{0}\) struggles to scale to large datasets when the ensemble size increases, we restrict our study to \model{FIPE}\(-\left\| \cdot \right\|_{1}\) and investigate its scalability to large ensembles. \cref{tab:ab-l1} shows that \model{FIPE}\(-\left\| \cdot \right\|_{1}\) is consistently able to reduce the number of active estimators while maintaining the fidelity of the original ensemble. Further, the results show that it scales well to large datasets and ensemble sizes. Another remarkable result shown by \cref{tab:ab-l1} is that the number of oracle calls $\noracle$ is relatively small, even for large ensembles. Indeed, it is possible to certify that the pruned model is functionally identical to the original one on the entire feature space with less than \(200\) calls.
\begin{table*}[hbt]
    \centering
    \small
    \input{tables/ab/ab-l1}
    \caption{Pruning with \model{FIPE}\(-\left\| \cdot \right\|_{1}\) on \model{AdaBoost} ensembles with \(100\), \(200\), \(500\), and \(1000\) base learners.}
    \label{tab:ab-l1}
\end{table*}

\begin{result}
    \model{FIPE}\(-\left\| \cdot \right\|_{1}\) consistently provides small pruned ensembles even when the dataset and original ensemble are large.
\end{result}

\begin{figure}[ht]
    \centering
    \makeatletter%
    \if@twocolumn%
        \setlength{\figurewidthx}{0.95\linewidth}%
    \fi
    \makeatother%
    \includegraphics[width=\figurewidthx]{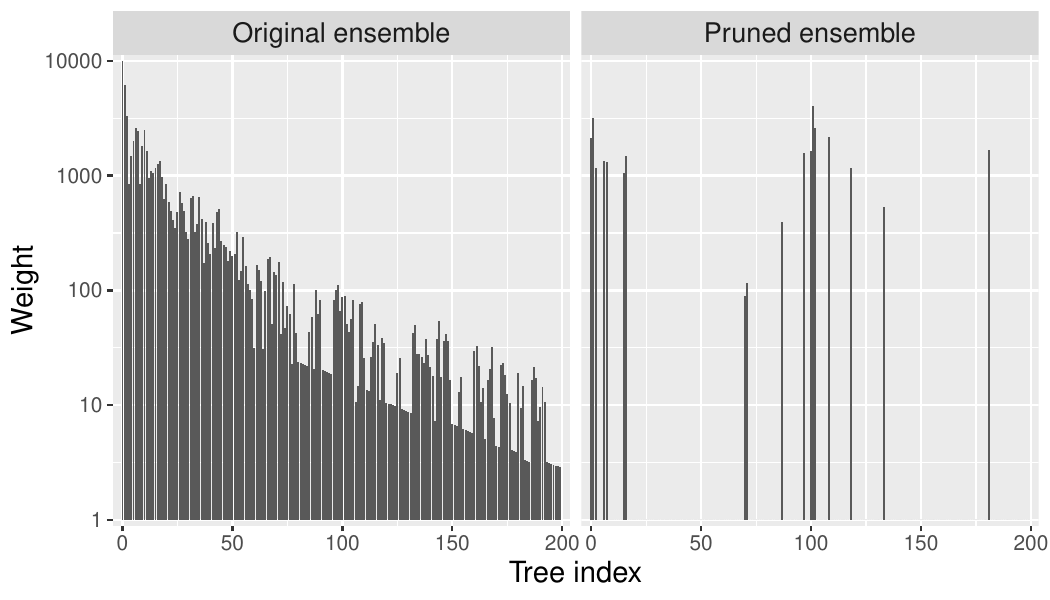}
    \caption{Weights of learners in the original and pruned ensembles on the {\sf FICO} dataset with \(M=200\).}
    \label{fig:FICO-200-weights}
\end{figure}
\subsection{Analysis of the pruned ensembles}
Our previous experiment shows that close to \(90\%\) of the base learners of \model{AdaBoost} ensembles are superfluous. That is, these learners can be effectively pruned without any change in the prediction function, if the remaining learners are reweighted adequately. We now study how \model{FIPE} is able to prune by jointly removing and reweighting the base learners. \cref{fig:FICO-200-weights} shows the weights of the active estimators in the original and pruned ensembles. It shows that some learners that seem to have a low predictive power in the original ensemble because they are created later in the training process (\cref{fig:FICO-200-weights} left) in fact have a large predictive power after pruning and reweighting (\cref{fig:FICO-200-weights} right). For instance, tree index 180 has a weight of around 30 in the original ensemble and a weight of around 1300 in the pruned model. Hence, it is possible to prune learners created early in the training process by assigning a large weight to later learners.

It is also interesting to observe that the size of the pruned ensemble grows significantly slower than the size of the original ensemble. This can be seen in \cref{tab:ab-l1}, which shows the size of the pruned ensemble for varying sizes of the original ensemble. Yet, the test accuracy tends to increase with the size of the ensemble. This suggests that only a small percentage of the newly created learners have a significant predictive power.


\begin{result}
    As boosted ensembles grow, their accuracy increases but they tend to have more superfluous base learners.
\end{result}

\subsection{Comparison with ``non-faithful'' baselines}
We compare \model{FIPE} with several baselines from other recent studies. To the best of our knowledge, we are the first to study functionally-identical pruning of ensembles. Hence, all existing baselines cannot guarantee faithfulness to the original ensemble. We implement the method of \citet{Guo2018margin} denoted as \model{IMD}, \citet{Lu2010ensemble} denoted as \model{IC} and \citet{Li2012diversity} denoted as \model{DREP}. We use the implementation from the \package{pypruning} package \citep{Buschjager2021improving}. A key difference with our approach is that these methods require the size of the pruned ensemble as a hyperparameter. To ensure a fair comparison, we set the value of this hyperparameter as the size of the pruned ensemble identified by \model{FIPE}.

\cref{tab:baselines} shows the faithfulness \(\faithmetric\) of the pruned model to the original model as well as the test accuracy \(\accuracy\) of the pruned model. The results show that the baselines have significantly lower accuracy and faithfulness to the original model on average, regardless of the size of the original ensemble. Hence, \model{FIPE} outperforms all baselines w.r.t. both faithfulness and accuracy.
\begin{table}[ht]
    \centering
    \small
    \input{tables/baselines}
    \caption{Comparison of \model{FIPE}\(-\left\| \cdot \right\|_{1}\) and non-faithful baselines for varying size of the original ensemble \(M\). Results are averaged over all datasets and repetitions.}
    \label{tab:baselines}
\end{table}

\begin{result}
    Functionally identical pruning guarantees that there is no decrease in test accuracy. In contrast, existing pruning methods have no guarantee and decrease the test accuracy in practice.
\end{result}

%% file: tables/ab/ab-l0-vs-l1.tex
\showLzeroVSLoneSingle{csv/ab/ab-l0-vs-l1.csv}

%% file: tables/datasets.tex
\showDatasets{csv/datasets.csv}

%% file: tables/ab/ab-l1.tex
\showLone{csv/ab/ab-l1.csv}{100}{200}
\showLone{csv/ab/ab-l1.csv}{500}{1000}

%% file: tables/baselines.tex
\showBaseline{csv/baselines-ab.csv}

%% file: sections/6-literature.tex
\section{Related Literature}
Many approaches have been proposed for ensemble pruning. We provide an overview of the closely related literature in this section, and refer an interested reader to \citet{Tsoumakas2009ensemble} and Chapter~6 of \citet{Zhou2012ensemble} for a more exhaustive review.

\paragraph{Pruning tree ensembles.} We can distinguish three main frameworks for pruning: order-based, cluster-based, and optimization-based. The first two are approximate and therefore do not guarantee that the pruned ensembles are of minimal size. Early approaches rank the elements of the ensemble and select them in a greedy fashion \citep{Margineantu1997pruning}. This can be regularized by encouraging the diversity of the selected classifiers \citep{Lu2010ensemble, Guo2018margin}. Cluster-based pruning is performed in two steps: first, cluster the classifiers in representative groups, then select a representative element from these groups \citep[see e.g.][]{Giacinto2000design}.

A few optimization-based approaches have also been proposed. They are based on mixed-integer quadratic formulations, which typically do not scale to large models and datasets. \citet{Zhang2006ensemble} present a quadratic formulation and propose a relaxation based on semi-definite programming. \citet{Li2012diversity} and \citet{Cavalcanti2016combining} extend this idea with diversity measures.

Finally, an interesting stream of literature focuses on building models with small memory requirements \citep{Kumar2017resource}. \citet{Painsky2019lossless} study how to compress random forests by encoding the trees into binary codes. \citet{Nakamura2022simplification} simplify tree ensembles by reducing the number of distinct splitting rules and sharing the rules across trees.

\paragraph{Optimal pruning.} Optimal methodologies have demonstrated effectiveness across several pruning tasks. \citet{Sherali2009optimal} present an optimal algorithm for pruning nodes of decision trees. \citet{Liu2023forestprune} show how to jointly prune nodes of random forest ensembles using a coordinated-descent algorithm. The tree-based structure of tree ensembles is especially suitable for combinatorial optimization methods, allowing to adapt the training algorithm or post-process ensembles to improve fairness or interpretability \citep{carrizosa2021mathematical}. Combinatorial optimization methods have also proven successful for pruning large deep neural networks \citep{Serra2020lossless, Yu2022combinatorial, Benbaki2023fast, Meng2024}.

\paragraph{Faithfulness.} Functionally identical pruning introduces a significant shift in perspective compared to existing works. Traditionally, the question asked was \quoteIt{given a target size, what is the best subset I should select from my ensemble?}. In contrast, we look for the smallest reweighted subset of trees that provide the same prediction function. The closest work to ours is likely \citet{Vidal2020born}, who show how to transform a tree ensemble into a decision tree with an identical prediction function on the entire feature space. However, this born-again tree is computationally expensive to obtain and might grow exponentially large in some cases as some ensembles are not efficiently representable as single trees.

%% file: sections/7-conclusion.tex
\section{Conclusion}
This paper proposed new methodological tools to prune additive tree ensembles while guaranteeing that their prediction function remains unchanged. Through extensive experimental analyses, we have demonstrated that boosted ensembles can be significantly compressed. This suggests that boosted ensembles have many superfluous learners, which can be removed entirely from the ensemble on the condition that the remaining learners are adequately reweighted. A significant advantage of such pruning is that, contrary to existing pruning methods, it guarantees that the test accuracy of the pruned ensemble does not decrease.

This work opens numerous research perspectives. First, while we focused on being certifiably identical to the original model on the entire feature space, it is worthwhile to study how to impose faithfulness on a subspace, such as the space of plausible observations as \citet{Parmentier2021}. This might allow stronger compression. For most practical situations, it is possible to avoid using the combinatorial optimization oracle. Using an approximate oracle may lead to a faster algorithm that lacks the faithfulness certificate but might empirically be very close to the original model. Last but not least, pursuing a disciplined analysis of model compression with faithfulness guarantees for other machine learning models is a promising direction for future work.

%% file: sections/8-acknowledgements.tex
\section*{Acknowledgements}

This work was supported by the Canada Excellence Research Chair in Data Science for Real-Time Decision-Making and the SCALE-AI Research Chair in Data-Driven Supply Chains. The authors would like to thank the anonymous reviewers for their valuable suggestions.

%% file: sections/a-method.tex
\section{Supplementary material: Method and implementation}
\label{app:method}
\subsection{Proofs}

\begin{proof}[Proof of \cref{prop:complexity}]
    \input{proofs/fipe-np-hard}
\end{proof}

\begin{proof}[Proof of \cref{thm:fipe-terminates}]
    \input{proofs/fipe-terminates}
\end{proof}

\subsection{Feature consistency for additive tree ensembles}
This section provides the additional constraints required in the separation oracle to ensure that the feature values are consistent with all the branching decisions. The formulation is taken from \citet{Parmentier2021}.

\paragraph{Binary features.} Let \(j\) be a binary feature. We assume that a split on the feature \(j\) is such that the tree branches to the left if the value of \(x_j\) is \(0\) and branches to the right otherwise. We consider the binary variable \(x_j\) that indicates if the value of \(x_j\) is \(1\). For each tree \(m\), let \(\nodes_{m}^{j}\) be the set of nodes that split on binary feature \(j\). We ensure that the value of \(x_j\) is consistent across the nodes of tree \(m\) with the constraints:
\begin{subequations}%
    \begin{alignat}{3}%
        &\flowvar_{m, \treeleft(\node)} &&\le 1-x_j,\; && \forall \node \in \nodes_{m}^{j}, \label{eq:sep-pair-cons-lin-binary-left}\\
        &\flowvar_{m, \treeright(\node)} &&\le x_j,\; && \forall \node \in \nodes_{m}^{j}. \label{eq:sep-pair-cons-lin-binary-right}
    \end{alignat}%
\end{subequations}%
Constraint \eqref{eq:sep-pair-cons-lin-binary-left} ensures that the value of \(x_j\) is \(1\) if we cannot go to the left child at node \(\node\). Similarly, Constraint \eqref{eq:sep-pair-cons-lin-binary-right} ensures that the value of \(x_j\) is \(0\) if we cannot go to the right child at node \(\node\).

\paragraph{Categorical features.}%
Let \(j\) be a categorical feature. We assume that the categories of the feature \(j\) are given by \(\left\{1, 2, \ldots, Z_j\right\}\). We introduce the binary variable \(\nu_{j, z}\) to indicates whether the value of \(x_j\) is \(z\) for categorical feature \(j\) for \(z\in \left\llbracket 1, Z_j \right\rrbracket\). First, we ensure that only a single category is active for a categorical feature using the constraint:
\begin{alignat}{2}%
    \sum_{z=1}^{Z_j}\nu_{j, z} &= 1. \label{eq:sep-pair-cons-lin-categorical-sum}%
\end{alignat}%
Constraint~\eqref{eq:sep-pair-cons-lin-categorical-sum} ensures that the value of \(x_j\) is only one of the categories. For each tree \(m\), let \(\smash{\nodes_{m}^{j, z}}\) be the set of nodes that split on categorical feature \(j\) and branch to the right child if the value of \(x_j\) is \(z\). We ensure that the value of \(\nu_{j, z}\) is consistent across the nodes of tree \(m\) with the constraints:
\begin{subequations}%
    \begin{alignat}{3}%
        &\flowvar_{m, \treeleft(\node)} &&\le 1-\nu_{j, z},\; && \forall z \in \left\llbracket 1, Z_j \right\rrbracket, \forall \node \in \nodes_{m}^{j, z}, \label{eq:sep-pair-cons-lin-categorical-left} \\
        &\flowvar_{m, \treeright(\node)} &&\le \nu_{j, z},\; && \forall z \in \left\llbracket 1, Z_j \right\rrbracket, \forall \node \in \nodes_{m}^{j, z}. \label{eq:sep-pair-cons-lin-categorical-right}%
    \end{alignat}%
\end{subequations}%
Constraint \eqref{eq:sep-pair-cons-lin-categorical-left} ensures that, if the value of \(x_j\) is \(z\), the path of point \(x\) in tree \(m\) at node \(\node\) does not go to the left child. Similarly, constraint \eqref{eq:sep-pair-cons-lin-categorical-right} ensures that if the value of \(x_j\) is not \(z\), the path of point \(x\) in tree \(m\) at node \(\node\) goes to the right child.

\clearpage

%% file: proofs/fipe-np-hard.tex
\citet{Vidal2020born} have demonstrated how to transform a 3-SAT instance into an additive tree ensemble that predicts \textsc{False} in the entire feature space if 3-SAT is \textsc{False}, and otherwise has a non-trivial prediction function. Applying functionally identical pruning in the sense of \cref{def:faithful} to the forest henceforth leads to a compressed result with a single tree predicting \textsc{False} if 3-SAT is \textsc{False}, and to a different result otherwise. Given that 3-SAT is NP-complete, it follows that Problem~\eqref{opt:fipe} is NP-hard.

%% file: proofs/fipe-terminates.tex
We consider the ensemble of trees \(\mathcal{T}_{1}, \dots, \mathcal{T}_{M}\) that are trained on the same dataset. Each tree \( \mathcal{T}_{m} \) is a piecewise constant function that divides the input space into \( J_{m} \) disjoint regions, \( R_{m,1}, \dots, R_{m,J_{m}} \), with a constant prediction assigned to each region. This proves that the prediction function of any weighted ensemble will be identical across any region of the form \(\bigcap_{m=1}^{M} R_{m, j_{m}}\) for \( j_{m} \in \left\llbracket 1, J_{m} \right\rrbracket\). The oracle will provide a new point only when there is a region where the predictions of the original and pruned ensembles differ. Consequently, the pruning process will terminate after a finite number of oracle calls at most equal to \(\prod_{m=1}^{M} J_{m}\).

%% file: sections/b-experiments.tex
\section{Supplementary material: Computational experiments}
\label{app:experiments}

This section presents the results of applying \model{FIPE} to various additive tree ensembles. The analysis includes:
\setlist{nolistsep}
\begin{itemize}[leftmargin=!]
    \setlength{\itemsep}{1pt}
    \item gradient-boosted trees (\model{Gradient Boosting}) using the implementation from the \package{scikit-learn} package \citep{scikit-learn},
    \item random forests (\model{Random Forest}) also from the \package{scikit-learn} package,
    \item \model{LightGBM} using the implementation from \citet{Ke2017lightgbm}, and
    \item \model{XGBoost} using the implementation from \citet{Chen2016xgboost}.
\end{itemize}

All hyperparameters are kept at their default values, except for the maximum depth of the trees, which is set to one or three. The experimental settings and datasets are kept as in the experiments presented in \cref{sec:experiments}. First, we provide a summary of \model{FIPE}'s pruning effectiveness on these tree ensembles, followed by detailed results for each ensemble.

\subsection{Summary of Results: Pruning and Accuracy}

\cref{tab:overview} presents the relative size of the ensemble after pruning with \model{FIPE}\(-\left\| \cdot \right\|_{1}\) calculated as \(\left\|w\right\|_{0}/M\) along with the test accuracy averaged across all datasets. The results show that \model{FIPE} effectively prunes the ensemble in most cases. On average, \model{Random Forests} tend to be pruned more effectively by \model{FIPE}. There is no significant difference in ensemble size after pruning between the different boosted tree ensembles. Ensembles with a smaller maximum depth tend to have smaller relative pruned sizes. For a fixed maximum depth, the relative size of the pruned ensemble decreases as the number of base learners increases.

\begin{result}
    \model{FIPE} can consistently prune various types of tree ensembles. The extent of pruning is relatively insensitive to the training algorithm and depends mostly on the maximum depth of each tree.
\end{result}

\begin{table}[ht]
    \centering
    \small
    \input{tables/l1-agg}
    \caption{Relative size of the pruned ensemble using \model{FIPE}\(-\left\| \cdot \right\|_{1}\) and test accuracy for different tree ensembles and maximum tree depths.}
    \label{tab:overview}
\end{table}

\subsection{Detailed results}
We now present detailed results for the different tree ensembles. First, we compare the performance of \model{FIPE}\(-\left\| \cdot \right\|_{0}\) and \model{FIPE}\(-\left\| \cdot \right\|_{1}\) in \cref{tab:overview-l0-vs-l1} for varying maximum depth. The results are consistent with those presented in the main paper. That is, the amount of pruning of \model{FIPE}\(-\left\| \cdot \right\|_{1}\) is often equal to the one of \model{FIPE}\(-\left\| \cdot \right\|_{0}\) with only a fraction of the computation time. This is true for all the ensemble models and datasets. On average, the difference between \model{FIPE}\(-\left\| \cdot \right\|_{0}\) and \model{FIPE}\(-\left\| \cdot \right\|_{1}\) is larger as the maximum tree depth increases.

{
\def\LongTableCaption{%
    \caption{Comparison of \model{FIPE}\(-\left\| \cdot \right\|_{0}\) and \model{FIPE}\(-\left\| \cdot \right\|_{1}\) for pruning tree ensembles with \(\pgfmathprintnumber{\LZEROEST}\) base learners.}%
    \label{tab:overview-l0-vs-l1}%
}
\small
\input{tables/l0}
}

Finally, \cref{tab:overview-l1} analyzes the scalability of \model{FIPE}\(-\left\| \cdot \right\|_{1}\) to ensembles of size \(100\) and \(200\). Again, the results are consistent with those presented in \cref{sec:experiments}. The amount of pruning as the number of base learners increases and tends to decrease with the maximum depth of the trees. Additionally, the relative size of the pruned ensemble is heavily dependent on the dataset. For ensembles of size $M=200$, \model{FIPE}\(-\left\| \cdot \right\|_{1}\) can always prune the ensemble, even if the amount of pruning is relatively low for more complex datasets such as \textsf{ELEC2} or \textsf{House-16H}.

{
\def\LongTableCaption{%
    \caption{Pruning with \model{FIPE}\(-\left\| \cdot \right\|_{1}\) on tree ensembles with \(100\) and \(200\) base learners.}%
    \label{tab:overview-l1}%
}
\small
\input{tables/l1}
}

%% file: tables/l1-agg.tex
\showLoneAgg{csv/l1-agg.csv}{50}{100}{200}

%% file: tables/l0.tex
\clearpage

\showLzeroVSLoneAll{csv/l0-vs-l1.csv}{\LongTableCaption}

%% file: tables/l1.tex
\clearpage

\showLoneAll{csv/l1.csv}{100}{200}{\LongTableCaption}